%% file: main.tex
\documentclass{article}
\usepackage{spconf,amsmath,graphicx}

\usepackage{colortbl}
\usepackage{tabularx}
\usepackage{float}
\usepackage{multirow} 
\usepackage{makecell}
\usepackage{tcolorbox}
\usepackage[dvipsnames]{xcolor}
\usepackage{stfloats}
\usepackage{booktabs}
\usepackage{amsfonts}
\usepackage{hyperref}
\newcolumntype{M}[1]{>{\centering\arraybackslash}m{#1}}

\definecolor{light-red}{RGB}{210,83,83}
\definecolor{lightgray}{gray}{0.9}
\definecolor{nicegreen}{RGB}{0,150,0}
\definecolor{nicered}{RGB}{150,0,0} 
\newcommand{\upgreen}[1]{\textcolor{nicegreen}{$\uparrow$ (#1)}}

\title{Faithful Grounded Visual Reasoning via Learned Proxy-Tokens}
\name{Tom Hodemon \quad Mohamed Chaouch \quad Aboubacar Tuo \quad Angelique Loesch} 
\address{Université Paris-Saclay, CEA, List, F-91120, Palaiseau, France} 
\begin{document}
%\ninept
%
\maketitle

\begin{abstract}

Multimodal Large Language Models (MLLMs) have achieved remarkable success in Visual Question Answering (VQA), yet their "black-box" nature hinders deployment in critical domains. Grounded Visual Reasoning (GVR) approaches attempt to improve interpretability by explicitly couple textual rationales with visual grounding information, which are typically textual coordinates. This mechanism lacks a learnable semantic link to the visual features, often resulting in a semantic-spatial gap where the model hallucinates coordinates that do not correspond to image evidences.
In this work, we introduce Composer, a MLLM that leverages a novel visual grounding mechanism based on learned proxy-tokens to promote faithful interpretability. These discrete symbolic pointers explicitly index the image latent space, allowing the model to manipulate visual regions as addressable, semantically manipulable sets. To rigorously validate our novel grounding mechanism, we constructed ComposerGCoT, a dataset synthesized to enable holistic assessment of reasoning consistency and grounding accuracy. Experimental results indicate that Composer achieves performance parity with its coordinate-based counterpart in final answer accuracy, while improving visual grounding accuracy by +9.0 points. By demonstrating that discrete proxy-tokens capture spatial semantics more effectively than typical textual coordinates, we establish that visual grounding mechanisms with learnable semantic links represent a promising path toward trustworthy and reliable MLLMs.~\footnote{The model weights and dataset annotations are available at \url{https://github.com/CEA-LIST/Composer}.}

\end{abstract}


\section{Introduction}
\label{sec:introduction}

\begin{figure*}[ht]
    \vspace{-4mm}
    \centering
    \includegraphics[width=0.8\linewidth]{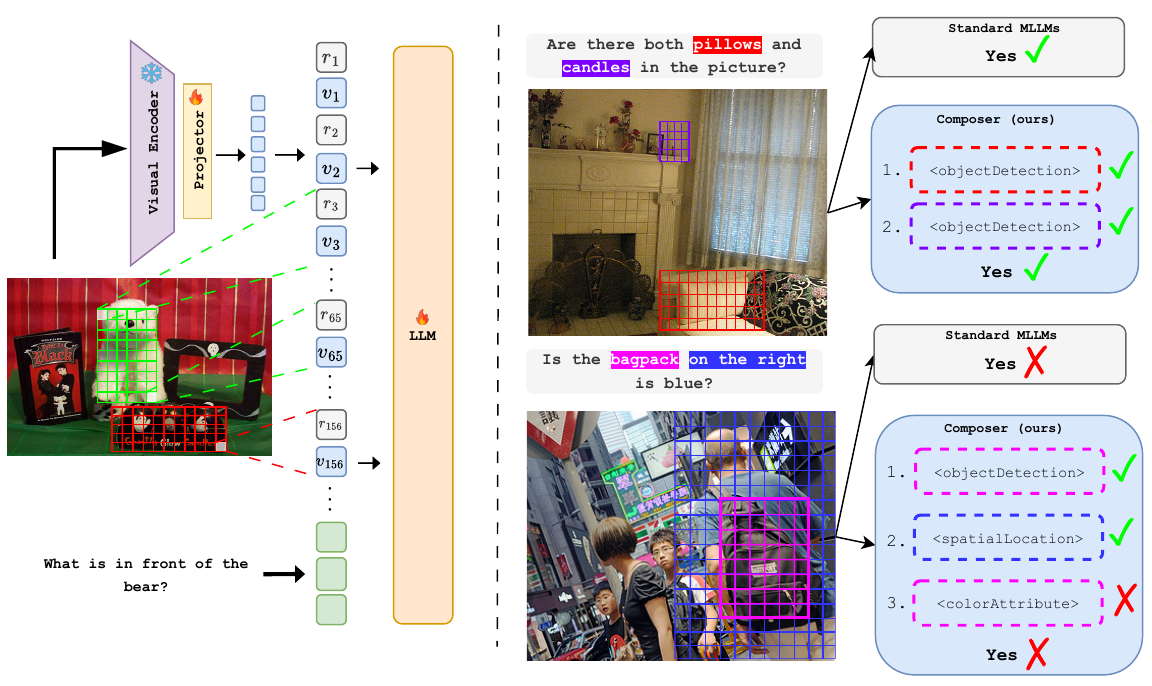}
    \caption{\textbf{(left) Visual Grounding via Proxy-Token Indexing.} We introduce a visual grounding mechanism that uses learned proxy-tokens to index the image latent space. Visual tokens $V = \{v_1, v_2, \dots, v_N\}$ are mapped to learnable proxy-tokens $R = \{r_1, r_2, \dots, r_N\}$ that serve as discrete pointers; the model references specific regions by addressing the proxy-tokens that span them. \textbf{(Right) Holistic Assessment.} We introduce Composer, an MLLM that leverages a novel visual grounding mechanism to facilitate faithfulness. Unlike standard approaches, Composer outputs structured reasoning chains that enable rigorous assessment across three axes: answer accuracy, reasoning consistency, and visual grounding accuracy.}
    \label{fig:overview}
\end{figure*}

Multimodal Large Language Models (MLLMs) have achieved remarkable success across diverse vision-language tasks~\cite{LLaVA, LLaVA-1.5, Blip-2}, most notably in Visual Question Answering (VQA). However, despite their impressive performance, these models operate under a monolithic paradigm that directly maps visual and textual inputs to a final answer without exposing their decision-making process. This lack of transparency presents a critical barrier to their deployment in critical applications.

To address this, recent research has focused on Grounded Visual Reasoning (GVR), aiming to equip MLLMs with the ability to provide interpretable, visually grounded rationales. Existing approaches typically adopt a coordinate regression paradigm, where the model is trained to output textual bounding box coordinates $(x, y, w, h)$ alongside its reasoning~\cite{GCoT, VISTAR, VoCoT}. While this offers a layer of interpretability, it suffers from a fundamental limitation: by treating spatial locations as arbitrary text strings, these methods lack a learnable semantic link to the visual features. This architectural disconnect creates a "semantic-spatial gap", where the model often hallucinates coordinates that are visually ungrounded, effectively decoupling the reasoning process from the image evidence.

In this work, we argue that faithful interpretability requires bridging this gap. We introduce Composer, an MLLM designed to enforce faithful, step-by-step grounded reasoning. At the core of our approach is a novel visual grounding mechanism based on learned proxy-tokens. Unlike traditional methods that use textual coordinates, Composer represents visual regions using discrete symbolic pointers that are interleaved with language tokens and explicitly index the image latent space (see Fig.~\ref{fig:overview}). This design promotes faithful visual grounding by allowing the model to manipulate visual regions as native, addressable elements of its vocabulary, thereby establishing a robust, learnable link between linguistic reasoning and visual perception.

However, a faithful model requires a rigorous evaluation. A major shortcoming in the current literature is that GVR methods are predominantly evaluated on final answer accuracy alone~\cite{Groma, vigorl, VoCoT}. This superficial protocol often masks the aforementioned disconnect between the reasoning process and the image evidence. To overcome this, we introduce ComposerGCoT, a specialized dataset synthesized to enable holistic assessment of model capabilities. By providing fine-grained supervision for every step of the reasoning chain, ComposerGCoT allows us to assess not just the final answer, but the consistency of the entire reasoning path and the fidelity of the visual grounding. 

Our contributions can be summarized as follows. We introduces Composer, a Multimodal Large Language Model (MLLM) that leverages a novel visual grounding mechanism based on learned proxy-tokens to promote faithful and precise grounded visual reasoning. To rigorously assess this approach, we present ComposerGCoT, a dataset and evaluation protocol designed to move beyond simple accuracy by enabling a holistic evaluation of reasoning consistency. This benchmark allows us to diagnose model reliability and determine if predictions are truly "right for the right reasons." Finally, through extensive experiments, we demonstrate that Composer achieves performance parity with its coordinate-based counterpart in final answer accuracy, while improving visual grounding accuracy by +9.0 points in IoU@0.95.

\section{Related Work}
\label{sec:related-work}

\textbf{Interpretability in the VQA Context.} Most existing MLLMs \cite{LLaVA, LLaVA-1.5, Blip-2} operate under a monolithic paradigm that directly maps visual and textual inputs to a final answer. Consequently, such models fail to expose their decision path, a critical deficiency for deployment in high-stakes applications~\cite{ke2025explainanswersurveycompositional}. 

To address this, several approaches have emerged. One line of research adapts Chain-of-Thought (CoT)~\cite{CoT} to the multimodal domain~\cite{LLaVA-CoT, MMCoT}. While these methods expose reasoning via textual rationales, they often lack explicit references to \textit{where} the model is looking. Conversely, tool-driven approaches~\cite{VISPROG, Vipergpt, wu2024mind, visual-sketchpad} leverage specialized visual modules~\cite{survey_thinking_with_images} to produce visual feedbacks. However, these systems suffer from high computational overhead and error propagation. Grounded Visual Reasoning (GVR)~\cite{VoCoT, GCoT, VISTAR, Visual-CoT, Groma, vigorl} is an other line of research. GVR approaches equip models with intrinsic grounding capabilities, enabling such models to explicitly couple textual rationales with visual grounding information. While promising, a limitation of existing approaches is that they often disregard the importance of faithful visual grounding by relying on simple coordinate-based mechanisms. Such mechanisms fail to establish a robust, learnable link between reasoning steps and visual evidence. Furthermore, current evaluation protocols predominantly assess performance based solely on final answer accuracy. This superficial evaluation masks cases where models rely on language biases or spurious correlations to guess the correct answer without performing the necessary visual reasoning steps. Our work addresses this gap by introducing a novel visual grounding mechanism designed to promote faithfulness, alongside a benchmark and evaluation protocol designed to assess reasoning capabilities beyond final answer accuracy.

\textbf{Visual Grounding Mechanisms.} A critical challenge in GVR lies in how the visual grounding information is represented within the language output. Methods such as VISTAR~\cite{VISTAR}, LLaVA-GCoT~\cite{GCoT}, and Volcano~\cite{VoCoT} treat localization as a regression task, outputting bounding box coordinates as "plain text" strings. While straightforward, this approach lacks a learnable semantic link to the visual features, often resulting in a "semantic-spatial gap" where the model hallucinates coordinates that do not correspond to image evidence. Our approach is more closely aligned with vision-centric models like Kosmos-2~\cite{Kosmos-2}, Florence-2~\cite{Florence2}, and Groma~\cite{Groma}, which discretize spatial locations into special vocabulary tokens. We extend this paradigm by explicitly indexing the image latent space with learned proxy-tokens, enabling an image entity to be represented as the set of proxy-tokens it spans.

\section{Method}
\label{sec:method}

In standard GVR approaches~\cite{GCoT, VISTAR, VoCoT} , the LLM processes spatial coordinates as arbitrary textual coordinates, which lack an explicit, learnable correspondence to the underlying image features. This creates a semantic-spatial gap: the model must reason about locations using linguistic symbols that are fundamentally ungrounded in the visual latent space. To bridge this gap, we introduce Composer, an GVR-capable model that leverages a novel visual grounding mechanism based on learned proxy tokens $\mathcal{R} = \{\texttt{<r}_1\texttt{>}, \dots, \texttt{<r}_N\texttt{>}\}$. Composer facilitates visual grounding through discrete symbolic pointers that directly index the image latent space. This design treats visual regions as addressable, semantically manipulable sets, aligning naturally with the Transformer’s native token-based architecture (see Fig.~\ref{fig:overview}).

\subsection{Visual Grounding via Learned Proxy-tokens}
Let an image be decomposed into a sequence of $N$ visual tokens $V = \{v_1, v_2, \dots, v_N\}$, where each $v_i \in \mathbb{R}^d$ represents a feature vector output by the multimodal projector. We define a corresponding set of $N$ learnable proxy-tokens $\mathcal{R} = \{\texttt{<r}_1\texttt{>}, \dots, \texttt{<r}_N\texttt{>}\}$, where each $\texttt{<r}_i\texttt{>}$ is a token added to the LLM's vocabulary. Thus, each proxy-tokens of $\mathcal{R}$ possess a token embedding $R = \{r_1, r_2, \dots, r_N\}$. The input sequence $\mathcal{S}$ is constructed by concatenating the interleaved sets of region tokens $\mathcal{R}$ and visual tokens $V$ with the user query $Q$:

\vspace{-3mm}
$$\mathcal{S} = [ r_1, v_1, r_2, v_2, r_3, v_3, \dots, r_N, v_N, Q ]$$
\vspace{-3mm}

%$$\mathcal{S} = [ \colorbox{Apricot}{\textcolor{white}{$r_1, v_1$}}, \colorbox{Emerald}{\textcolor{white}{$r_2, v_2$}}, \colorbox{LimeGreen}{\textcolor{white}{$r_3, v_3$}}, \dots, \colorbox{Periwinkle}{\textcolor{white}{$r_N, v_N$}}, Q ]$$

% \begin{tcolorbox}[
%     title={Input Sequence},
%     fonttitle=\bfseries,
%     % fontupper=\footnotesize,
%     colback=gray!10,
%     colframe=gray!70,
%     arc=8px,
%     boxrule=2pt,
%     left=5pt, right=5pt, top=5pt, bottom=5pt
% ]
%     % \raggedright is crucial for long token strings to avoid "Underfull \hbox" errors
%     \raggedright 
%     \textbf{User}: 
%     $\texttt{<r}_{1}\texttt{>}\allowbreak\texttt{<image>}\allowbreak
%     \texttt{<r}_{2}\texttt{>}\allowbreak\texttt{<image>}\allowbreak
%     \texttt{<r}_{3}\texttt{>}\allowbreak\texttt{<image>}\allowbreak
%     \texttt{<r}_{4}\texttt{>}\allowbreak\texttt{<image>}\allowbreak
%     \texttt{<r}_{5}\texttt{>}\allowbreak\texttt{<image>}\dots\allowbreak
%     \texttt{<r}_{256}\texttt{>}\allowbreak\texttt{<image>}. \texttt{\{QUESTION\}}$
% \end{tcolorbox}

With this arrangement, we explicitly index each $\texttt{<r}_i\texttt{>}$ with its corresponding visual token $v_i$, maintaining a strict 1-to-1 correspondence. During training, we teach the model to refer to visual regions addressing the corresponding learnable proxy-tokens. This novel visual grounding mechanism effectively transform the visual feature space into a set of "addressable memory slots", grounding the model’s reasoning directly in the physical structure of the image.

\begin{figure}[h]
    \centering
    \includegraphics[width=0.8\linewidth]{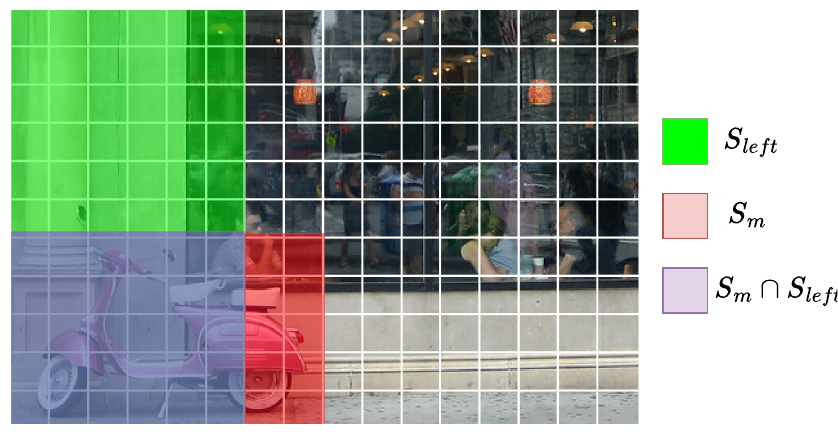}\caption{\textbf{Spatial reasoning via implicit set-based operation.} The visual entity "motorcycle" $S_m$ and the left-hand spatial prior $S_{left}$ are mapped to proxy-token subsets. The model evaluates spatial predicate "to the left of" by implicitly identifying non-empty intersections between these two sets: $S_m \cap S_{left} \neq \emptyset $.}
    \label{fig:method:proxy-tokens}
\end{figure}

This symbolic representation empowers the model to perform implicit set-based operations—such as intersection, union, and subset verification—directly on visual regions. A visual entity $\mathcal{O}$ is no longer defined by abstract bounding box coordinates, but by the discrete subset of proxy tokens it spans: $S_\mathcal{O} \subseteq \mathcal{R}$. By representing entities as proxy-token sets, the LLM can leverage its self-attention layers to map linguistic predicates (e.g., "inside", "overlapping", "left of") directly to set-based operations within the proxy-token set. For instance, as illustrated Fig.~\ref{fig:method:proxy-tokens}, to determine if a "motorcycle" is "to the left of the image," the model identifies the set of tokens for the motorcycle ($S_m$) and the set representing the left-hand spatial prior ($S_{left}$). The spatial reasoning task is thus reduced to a set-based intersection:

\vspace{-3mm}
$$\text{IsLeft}(m) \iff S_m \cap S_{left} \neq \emptyset$$
\vspace{-3mm}

Furthermore, this mechanism facilitates recursive reasoning: to infer an object's attribute (e.g., color), the model refers back to the proxy tokens generated in previous reasoning steps, effectively "shrinking" the visual search space to the relevant indices. We hypothesis that this transition from regression to discrete set manipulation promote a more spatially faithful and logically consistent grounded reasoning.

\subsection{Visually Grounded Annotations}
\label{subsec:construction-composergcot}

To train Composer, we employ a two-stage data strategy designed to first establish spatial alignment and subsequently cultivate multi-step grounded visual reasoning.

\textbf{Stage 1: Alignment Pretraining.}  We first equip the model with fine-grained localization capabilities by aggregating a diverse suite of vision-language tasks. To learn global image-text alignment, we utilize captioning data from CC3M~\cite{CC3M} and Flickr30K~\cite{flickr30k}. For region-level precision, we incorporate referring expression comprehension (REC) tasks from COCO~\cite{COCO}, RefCOCO/+/g~\cite{refcoco}, and Visual Genome~\cite{VisualGenome}. Finally, we include Visual Spatial Reasoning~\cite{VSR} to ensure a robust understanding of object relationships. For each sample, bounding box annotation is mapped to our proxy-token $\mathcal{R}$.

\textbf{Stage 2: Multi-step Reasoning via ComposerGCoT.} Standard multimodal datasets~\cite{GCoT, VISTAR} often lack the intermediate spatial supervision required for complex reasoning. To address this, we introduce \textbf{ComposerGCoT}, a specialized dataset synthesized by distilling GQA~\cite{GQA}. Leveraging GQA’s scene graphs and functional programs, we employ a rule-based pipeline to decompose SQL-like reasoning paths into discrete, visually grounded steps. Each step is annotated with intermediate answers and visual grounding annotations, resulting in 163K high-quality reasoning chains for 55K images. Examples of sample from ComposerGCoT can be found in Appendix~\ref{appendix:examples-composergcot}.

\subsection{Structured Output for Holistic Evaluation}
\label{sec:method:structured-output}

Evaluating the internal consistency of MLLM reasoning is often hindered by the unstructured nature of free-form text. To address this, Composer adopts a structured output format using lightweight XML-style tags that represent discrete visual capabilities—defined here as atomic visual-reasoning primitives (e.g., object detection, spatial filtering, or attribute recognition). The model encapsulates its reasoning process within a \verb|<think>| block, where it generates a sequence of these capability tags before producing the final \verb|<answer>|. We derive these sequences directly from the GQA dataset’s functional programs; for every question, the GQA operational logic (such as \textit{select}, \textit{filter}, or \textit{relate}) is mapped to its corresponding visual capability in our XML schema. Each tag encapsulates proxy-tokens that ground that specific reasoning step within the image. 
For instance, to answer \textit{"Is the dog on the right white?"}, the model decomposes the query into a sequential path of capabilities: first locating the subject (\verb|<objectRecognition>|), then applying a spatial constraint (\verb|<spatialLocation>|), and finally verifying the target property (\verb|<attributeColor>|). This design enables a holistic evaluation framework: beyond final accuracy, we can independently verify the spatial and logical validity at step-level. a full taxonomy of tags and corresponding generation examples, refer to Appendices~\ref{appendix:taxonomy-tags} and \ref{appendix:examples-generation}.

\begin{figure}
    \centering
    \includegraphics[width=\linewidth]{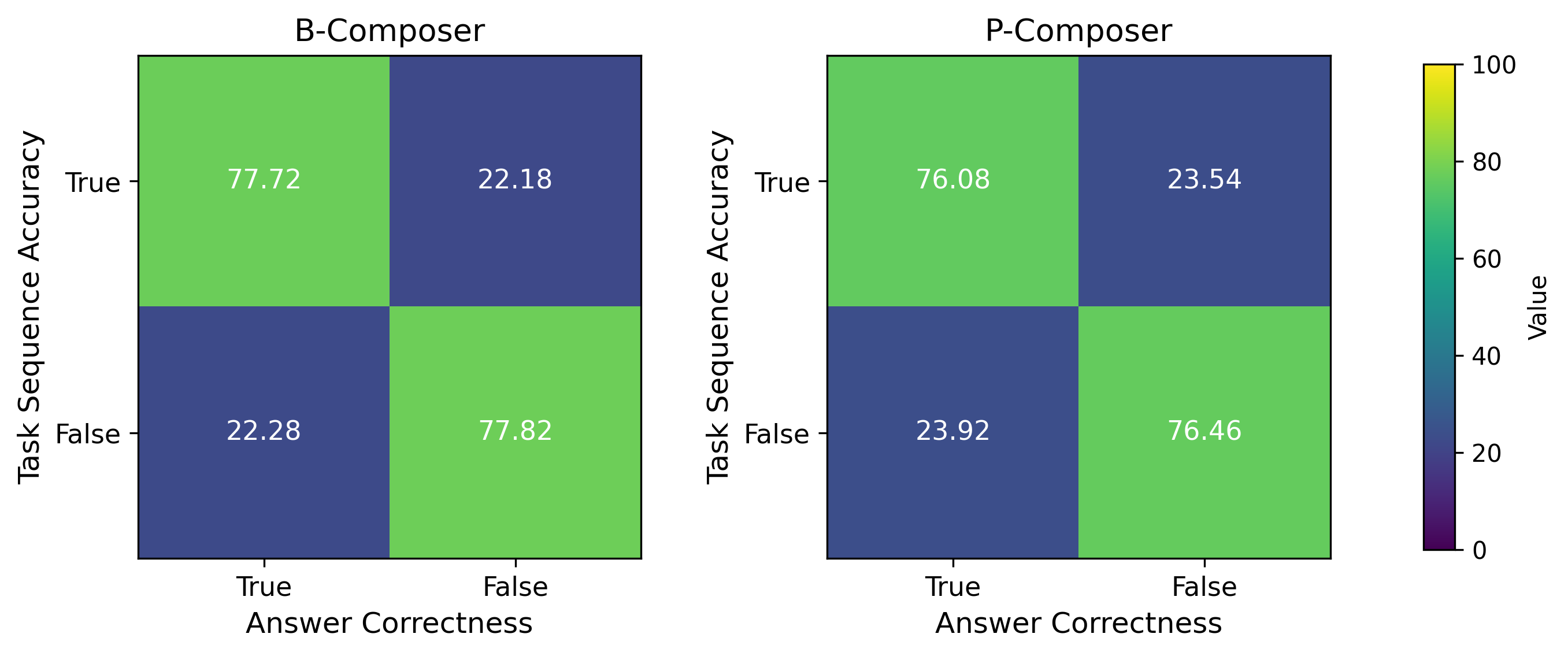}
    \caption{\textbf{Reasoning consistency.} These heatmaps illustrate the logical consistency of B-Composer and P-Composer by correlating task sequence accuracy (the model's ability to follow the ground-truth reasoning chain) with final answer correctness.}
    \label{fig:composergcot:heatmaps}
\end{figure}

\section{Experiments}
\label{sec:experiments}

\subsection{Implementation Details}
\label{subsec:experiments:implementation-details}

Composer follows the encoder-projector-LLM paradigm. We employ CLIP ViT-L/16~\cite{CLIP} as the visual encoder, coupled with a two-layer MLP projector that maps visual features into the embedding space of Vicuna-v1.5-7B~\cite{Vicuna}. To maintain a 1-to-1 mapping with the 256 visual tokens produced by CLIP ViT-L/16, we expanded the LLM vocabulary with 256 corresponding proxy tokens. Throughout the training process, the visual encoder remains frozen to preserve pretrained representations. Optimization is performed using AdamW across two stages using the next-token prediction objective. In the initial alignment stage (Stage 1), we train for 2 epochs with a learning rate of $2\text{e-}4$. In the subsequent stage (Stage 2), we fine-tune for 1 epoch with a reduced learning rate of $1\text{e-}5$. Comprehensive hyperparameters and infrastructure details are provided in Appendix~\ref{appendix:additional-training-details}.

\subsubsection{Benchmark and metrics} 

We evaluate on \textbf{ComposerGCoT-val}, a validation set comprising 3,959 samples across 2,943 images, synthesized from the GQA validation split using our pipeline described Sec.~\ref{subsec:construction-composergcot}. 

%Unlike standard VQA benchmarks~\cite{} that only check the final answer, this benchmark includes ground-truth reasoning traces (intermediate steps and bounding boxes), enabling us to assess the model's entire thought process.

Beyond simple accuracy, we assess the model along three complementary axes to provide a more holistic view of its performance. First, we evaluate \textbf{answer accuracy} to determine if the model predicts the correct final answer, which we calculate using exact string matching.To understand how the model reaches those conclusions, we then examine \textbf{reasoning consistency} via the task sequence accuracy. We leverage our structured output format to parse the generated sequence of visual capabilities produced by the model. We use exact match accuracy against the ground-truth reasoning chain to verify that the model follows a logical path. Finally, we measure \textbf{visual grounding accuracy} to ensure the visual rationales themselves are precise. For this, we assess spatial alignment using $IoU@k$ across the entire reasoning sequence; specifically for P-Composer, this $IoU$ is calculated by mapping proxy-token sets back into bounding boxes.

\subsubsection{Baselines}

To rigorously isolate the impact of our proposed learned proxy-token mechanism, we evaluate two distinct variants of the Composer framework. These variants share the exact same architecture, training budget, and are required to output structured output requirement (see Sec.~\ref{sec:method:structured-output}) differing only in their visual grounding mechanism:

\begin{enumerate}
    \item \textbf{B-Composer (Bounding Box):} This variant represents the standard regression-based approach. It treats spatial rationales as textual coordinate strings $[x, y, w, h]$, analogous to methods like GCoT~\cite{GCoT}.
    \item \textbf{P-Composer (Proxy-tokens):} This variant implements our novel visual grounding via learned proxy-tokens mechanism.
\end{enumerate}

% We compare final answer accuracy scores against LLaVA-1.5-7B~\cite{LLaVA-1.5}, which we refer to as LLaVA in the following section, a widely adopted MLLM. This serves as a reference point to verify that Composer maintains competitive general reasoning capabilities while offering enhanced interpretability in the reasoning process.

\subsection{Results and Analysis}
\label{subsec:evaluation:results-and-analysis}

\begin{table}[b]
    \centering
    \setlength{\tabcolsep}{5.5pt}
    \renewcommand{\arraystretch}{1.1}
    \footnotesize
    \begin{tabular}{l cc c cc c cc}
        \toprule
        & \multicolumn{2}{c}{\textbf{Correct}} && \multicolumn{2}{c}{\textbf{Incorrect}} && \multicolumn{2}{c}{\textbf{Overall}} \\
        \cmidrule(lr){2-3} \cmidrule(lr){5-6} \cmidrule(lr){8-9}
        \textbf{Metric} & B & P && B & P && B & P \\
        \midrule
        IoU        & 60.0 & \textbf{62.4} && 38.3 & \textbf{43.5} && 55.2 & \textbf{57.9} \tiny{\upgreen{2.7}} \\
        IoU@0.5    & 68.9 & \textbf{69.9} && 42.6 & \textbf{48.1} && 63.1 & \textbf{64.7} \tiny{\upgreen{1.6}} \\
        IoU@0.75   & 43.5 & \textbf{44.9} && 27.3 & \textbf{30.8} && 39.9 & \textbf{41.6} \tiny{\upgreen{1.7}} \\
        IoU@0.95   & 6.9  & \textbf{16.4} && 4.8  & \textbf{12.3} && 6.4  & \textbf{15.4} \tiny{\upgreen{9.0}} \\
        \bottomrule
    \end{tabular}
     \caption{\textbf{IoU@k scores by answer correctness.} Direct comparison of B-Composer (B) and P-Composer (P). P-Composer consistently outperforms B-Composer across all splits.}
     \label{table:grounding-accuracy}
\end{table}

\subsubsection{Reasoning consistency}
A risk in CoT reasoning is hallucination—where the model arrives at the correct answer despite a flawed reasoning path. While this phenomenon is well-documented, systematically diagnosing it has been challenging due to a lack of benchmarks with fine-grained intermediate supervision. \textbf{ComposerGCoT} addresses this limitation, enabling us to assess to go beyond the final answer accuracy.

Fig.~\ref{fig:composergcot:heatmaps} illustrates a strong correlation between task sequence and final answer accuracy. For P-Composer, a correct reasoning path leads to a correct answer in $76.08\%$ of cases, while flawed reasoning rarely yields correct predictions ($76.46\%$ False/False density). This causal link suggests that reasoning steps drive the final prediction rather than serving as post-hoc justifications. While B-Composer exhibits slightly higher True/True performance ($77.72\%$ vs. $76.08\%$), the gap likely stems from accumulated noise during P-Composer's extended proxy-token generation. However, this marginal trade-off is justified by the substantial improvements in grounding accuracy detailed in Section~\ref{subsec:grounding_results}.

\begin{table}
    \centering
    \setlength{\tabcolsep}{12pt}
    \renewcommand{\arraystretch}{1.1}
    \footnotesize
    \begin{tabular}{lc}
        \toprule
        \textbf{Visual Capability} & \textbf{Accuracy (\%)} \\
        \midrule
        objectRecognition   & 76.1 \\
        spatialRelationship & 79.2 \\
        attributeColor      & 66.5 \\
        spatialPosition     & 84.8 \\
        \midrule
        \textbf{Overall}    & \textbf{76.3} \\
        \bottomrule
    \end{tabular}
    \caption{\textbf{P-Composer Answer Accuracy by Visual Capabilities.} The model shows the highest performance in spatial positioning and lowest in color attribution.}
    \label{table:p-composer-accuracy}

    \vspace{-5mm}
\end{table}

\subsubsection{Learned proxy-tokens on grounding accuracy}
\label{subsec:grounding_results}

Our central hypothesis is that learned proxy-tokens (P-Composer) provides better grounding than coordinate regression (B-Composer). The results in Table~\ref{table:grounding-accuracy} validate this. P-Composer consistently outperforms the bounding-box variant in localization precision. Specifically, the proxy token mechanism improves overall IoU by 2.7 points and achieves a significant 9.0 points gain in strict IoU@0.95. This confirms that establishing an explicit, learned link between vocabulary tokens and visual features enables significantly sharper and more faithful visual grounding than treating coordinates as arbitrary text strings.

\subsubsection{Performances per Visual Capabilities}
\label{subsubsec:perf-per-cap}

In Table~\ref{table:p-composer-accuracy}, we report the answer accuracy for questions requiring specific reasoning steps. This detailed breakdown is made possible by the richness of our annotations, which allow us to isolate and evaluate specific visual capabilities demands. Each score represents the model's accuracy on the subset of questions that necessitate at least one operation of the corresponding visual capability (e.g., objectRecognition, spatialLocation). The model achieves its highest accuracy in spatialPosition ($84.8\%$) and spatialRelationship ($79.2\%$), while performance is lowest in attributeColor ($66.5\%$). We hypothesize that this discrepancy stems from the nature of the expected outputs: while spatial and recognition tasks often involve constrained choices or binary (yes/no) answers, color attribution frequently requires open-ended responses, which naturally increases the difficulty of the task.

\vspace{-4mm}
\section{Conclusion}
\label{sec:conclusion}

In this work, we introduce Composer, a GVR-capable model that bridges the gap between reasoning accuracy and faithful interpretability. Central to our approach is a novel visual grounding mechanism based on learned proxy-tokens, which establishes an explicit link between linguistic reasoning and visual features. To rigorously validate this paradigm, we constructed ComposerGCoT, a specialized dataset enabling holistic assessment of reasoning consistency and grounding faithfulness. Our experiments demonstrate that Composer maintains answer accuracy while significantly outperforming coordinate-based variant in visual grounding precision, leading to a more faithful grounded reasoning. By proving that discrete learned proxy-tokens capture spatial semantics more effectively than continuous coordinates, we demonstrate that establishing explicit, learnable links between modalities constitutes a promising direction for future research in trustworthy and interpretable MLLMs.

\textbf{Acknowledgments.} This publication was made possible by the use of the FactoryIA supercomputer, financially supported by the Ile-de-France Regional Council.
\vspace{-4mm}
\bibliographystyle{IEEEbib}
\bibliography{refs}

\appendix

\clearpage 

\input{sup_mat.tex} 

\end{document}

%% file: sup_mat.tex
\section{Additional Training Details}
\label{appendix:additional-training-details}

Table~\ref{table:appendix:training_details:hp} summarizes the detailed hyperparameters used for training both Composer variants. Table~\ref{table:appendix:training_details:durations} details the training durations for Composer variants. Both variants were trained on 4 H100 GPUs. For some large-scale datasets, we trained on a sampled subset. The total number of training samples per epoch is reported in Table~\ref{table:appendix:training_details:hp}.

\vspace{6mm}

\begin{table}[ht]
    \centering
    {
    \setlength{\tabcolsep}{12pt}
    \renewcommand{\arraystretch}{1.3}
    \begin{tabularx}{0.9\linewidth}{c *{2}{>{\centering\arraybackslash}X}}
    \toprule
    Variant & Stage 1 & Stage 2 \\
    \midrule
    Bounding Boxes & 1.5 days & 1 hour \\
    Proxy Tokens & 2 days & 1.5 hour \\
    \bottomrule
    \end{tabularx}
    }
    \caption{Detailed training durations for Composer variants.}
    \label{table:appendix:training_details:durations}
\end{table}

\vspace{6mm}
\begin{table}[ht]
    \centering
    {
    \setlength{\tabcolsep}{2pt}
    \renewcommand{\arraystretch}{1.3}
    \begin{tabularx}{0.9\linewidth}{c *{2}{>{\centering\arraybackslash}X}}
    \toprule
    Configuration & Stage 1 & Stage 2 \\
    \midrule
    optimizer & AdamW & AdamW \\
    epochs & 2 & 1 \\
    batch size & 128 & 128 \\
    learning rate & 2e-4 & 1e-5 \\
    weight decay & 0 & 0 \\
    training samples & 2.4m & 163k \\
    trainable param. & MLP, LLM & MLP, LLM \\
    \bottomrule
    \end{tabularx}
    }
    \caption{Hyperparameters used to train Composer variants.}
    \label{table:appendix:training_details:hp}
\end{table}

\section{Examples of ComposerGCoT Annotations}
\label{appendix:examples-composergcot}

\begin{figure}[ht]
  \centering
  \begin{minipage}{1.0\columnwidth}
    \centering
    \includegraphics[width=\linewidth]{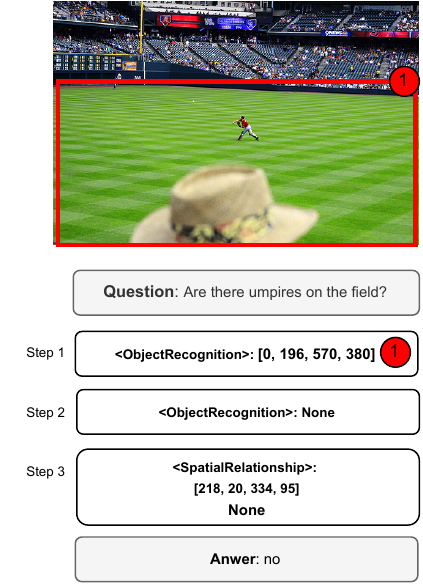}
    \caption{Example when an object mentioned in the question does not exist in the image.}
    \label{appendix:example_3}
  \end{minipage}
  
\end{figure}

\begin{figure*}
    \centering
    \includegraphics[width=\textwidth]{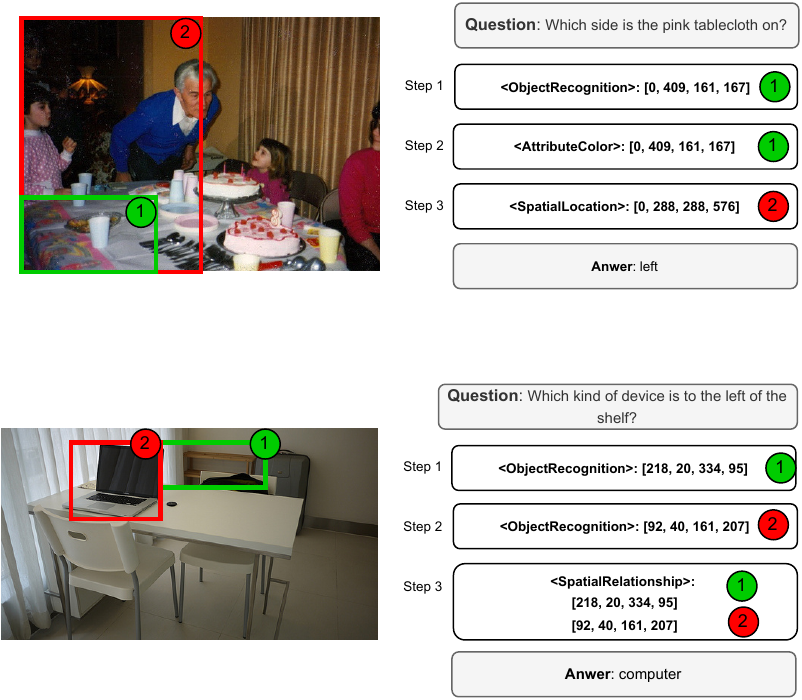}
    \caption{Each triplet consists of an image, question, and answer, annotated with a sequence of visual capabilities. These steps represent a consistent logical path to the final answer, with each individual step visually grounded in the image.}
    \centering
    \label{appendix:example_4}
\end{figure*}

Fig.~\ref{appendix:example_3} and Fig.~\ref{appendix:example_4} present samples from our ComposerGCoT dataset. 

\section{Examples for P-Composer Generation}
\label{appendix:examples-generation}

\begin{figure*}[t]
    \centering
    \includegraphics[width=\textwidth]{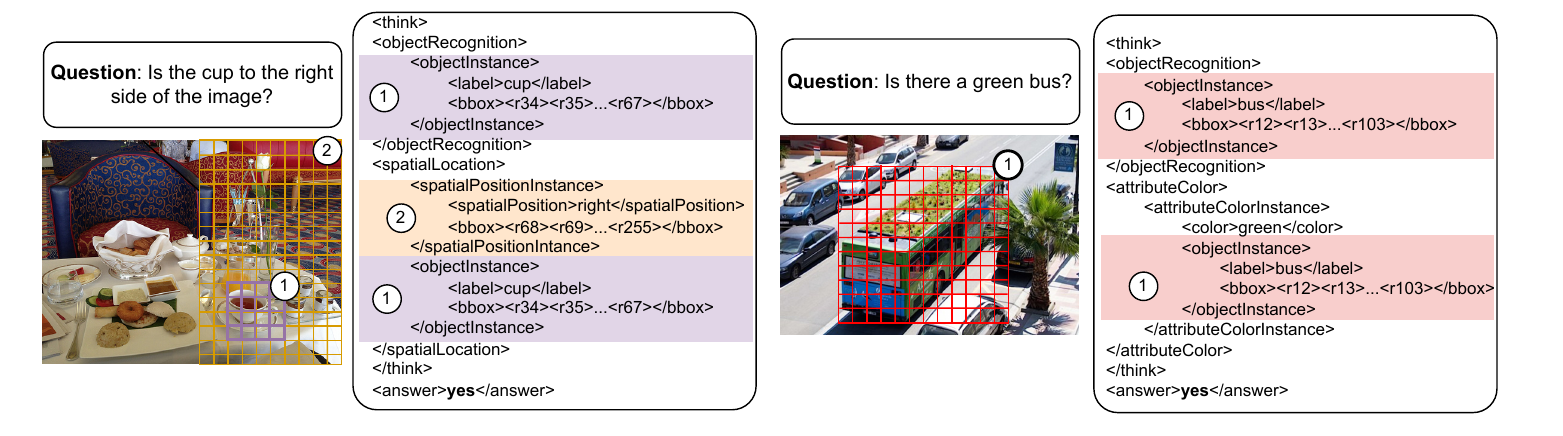}
    \caption{Examples of generation.}
    \label{fig:example-generation}
\end{figure*}

Fig.~\ref{fig:example-generation} illustrates the sequences generated by P-Composer (proxy-tokens variant).

\clearpage
\section{Taxonomy of tags}
\label{appendix:taxonomy-tags}

The Table~\ref{table:appendix:format:tags} presents the set of tags used to structure the ouput of Composer. 

\begin{table*}
    \centering
    \footnotesize
    {
    \setlength{\tabcolsep}{2pt}
    \renewcommand{\arraystretch}{1}
    \begin{tabularx}{\linewidth}{>{\centering\arraybackslash}p{3cm} >{\centering\arraybackslash}p{8cm} >{\centering\arraybackslash}p{3.5cm}}
    \toprule
    \textbf{Type} & \textbf{Description} & \textbf{Special Token} \\
    \midrule
    \multirow{4}{*}{Default}
        & \multirow{2}{=}{Marks the beginning and the end of the reasoning sequence} & \texttt{<reasoning>} \\
        & & \texttt{</think>} \\
        & \multirow{2}{=}{Marks the beginning and the end of the answer sequence} & \texttt{<answer>} \\
        & & \texttt{</answer>} \\
    \midrule
    \multirow{8}{*}{Visual Capability}
        & \multirow{2}{=}{Marks the use of object recognition capability} & \texttt{<objectRecognition>} \\
        & & \texttt{</objectRecognition>} \\
        & \multirow{2}{=}{Marks the use of spatial position capability} & \texttt{<spatialPosition>} \\
        & & \texttt{</spatialPosition>} \\
        & \multirow{2}{=}{Marks the use of attribution capability (color specialized)} & \texttt{<attributeColor>} \\
        & & \texttt{</attributeColor>} \\
        & \multirow{2}{=}{Marks the use of spatial relationship capability} & \texttt{<spatialRelationship>} \\
        & & \texttt{</spatialRelationship>} \\
    \midrule
    \multirow{10}{*}{Instance}
        & \multirow{2}{=}{Represents a detected object instance} & \texttt{<objectInstance>} \\
        & & \texttt{</objectInstance>} \\
        & \multirow{2}{=}{Same as \texttt{objectInstance}} & \texttt{<subjectInstance>} \\
        & & \texttt{</subjectInstance>} \\
        & \multirow{2}{=}{Represents a relation instance} & \texttt{<relationInstance>} \\
        & & \texttt{</relationInstance>} \\
        & \multirow{2}{=}{Represents a spatial position instance} & \texttt{<spatialPositionInstance>} \\
        & & \texttt{</spatialPositionInstance>} \\
        & \multirow{2}{=}{Represents an attribution instance (color specialized)} & \texttt{<attributeColorInstance>} \\
        & & \texttt{</attributeColorInstance>} \\
    \midrule
    \multirow{6}{*}{\shortstack{Label or \\ intermediate answer}}
        & \multirow{2}{=}{Object name label} & \texttt{<label>} \\
        & & \texttt{</label>} \\
        & \multirow{2}{=}{Color name label} & \texttt{<color>} \\
        & & \texttt{</color>} \\
        & \multirow{2}{=}{Relation name label, position of the object in the image} & \texttt{<relation>} \\
        & & \texttt{</relation>} \\
    \midrule
    \multirow{2}{*}{Visual Justification}
        & \multirow{2}{=}{Contains a location representation, serving as visual justification} & \texttt{<bbox>} \\
        & & \texttt{</bbox>} \\
    \bottomrule
    \end{tabularx}
    }
    \caption{\textbf{Taxonomy of tags.}}
    \label{table:appendix:format:tags}

\end{table*}